\documentclass[journal]{IEEEtran}
\usepackage{amsmath,amsfonts}
\usepackage{ragged2e}
\usepackage{algorithmic}
\usepackage{algorithm}
\usepackage{array}
\usepackage{textcomp}
\usepackage{stfloats}
\usepackage{url}
\usepackage{verbatim}
\usepackage{graphicx}
\usepackage{cite}
\usepackage[utf8]{inputenc}
\usepackage{booktabs}
\usepackage[caption=false,font=footnotesize]{subfig}
\begin{document}

\title{Quantum Machine Learning Approaches for Coordinated Stealth Attack Detection in Distributed Generation Systems}

\author{Osasumwen Cedric Ogiesoba-Eguakun,~\IEEEmembership{Member,~IEEE,} Suman Rath,~\IEEEmembership{Member,~IEEE}
\thanks{Manuscript submitted: December 30, 2025. This work was conducted as part of the graduate research activities at the University of Tulsa.\\ 
(Corresponding author: Osasumwen Cedric Ogiesoba-Eguakun.)
 
O. C. Ogiesoba-Eguakun and Suman Rath are with the Department of Electrical and Computer Engineering, The University of Tulsa, Tulsa, OK 74104, USA (e-mail: oco1411@utulsa.edu, suman-rath@utulsa.edu).}

}



\maketitle

\begin{abstract}
Coordinated stealth attacks are a serious cybersecurity threat to distributed generation systems because they modify control and measurement signals while remaining close to normal behavior, making them difficult to detect using standard intrusion detection methods. This study investigates quantum machine learning approaches for detecting coordinated stealth attacks on a distributed generation unit in a microgrid. High-quality simulated measurements were used to create a balanced binary classification dataset using three features: reactive power at DG1, frequency deviation relative to the nominal value, and terminal voltage magnitude. Classical machine-learning baselines, fully quantum variational classifiers, and hybrid quantum–classical models were evaluated. The results show that a hybrid quantum–classical model combining quantum feature embeddings with a classical RBF support vector machine achieves the best overall performance on this low-dimensional dataset,  with a modest improvement in accuracy and F1 score over a strong classical SVM baseline. Fully quantum models perform worse due to training instability and limitations of current NISQ hardware. In contrast, hybrid models train more reliably and demonstrate that quantum feature mapping can enhance intrusion detection even when fully quantum learning is not yet practical.

\end{abstract}


\section{Introduction}
\IEEEPARstart{T}{he} increasing integration of distributed generation (DG) in modern power systems has improved operational flexibility and grid resilience, but has also introduced new cyber–physical vulnerabilities through expanded communication, control, and supervisory infrastructures \cite{katiraei2008microgrids,hatziargyriou2014microgrids, ogiesoba2023design}. These interfaces can be exploited to launch coordinated stealth attacks that manipulate control or measurement signals while maintaining measurements close to nominal operating ranges \cite{lopes2006defining,lasseter2002microgrids}. Because such attacks are designed to mimic normal physical behavior, they are difficult to detect and often evade traditional intrusion detection methods \cite{goes2017stealthy,xie2011integrity}.
Machine-learning techniques have been widely applied to power-system cybersecurity and have demonstrated strong performance in detecting disturbances and cyberattacks using supervised learning and pattern recognition \cite{mukherjee2022deep,vijayanand2017support}. Classical approaches, including logistic regression, support vector machines (SVM) \cite{schuld2019quantum}, and ensemble techniques, remain strong baseline methods for intrusion detection \cite{li2022detection}. However, coordinated stealth attacks introduce subtle and nonlinear perturbations in voltage, reactive power, and frequency that lie close to normal operating manifolds, making robust class separation challenging for many classical algorithms \cite{teixeira2010cyber,bishop2006pattern}.
Quantum machine learning (QML) has recently emerged as a promising alternative for complex classification problems \cite{schuld2015introduction}. By exploiting quantum superposition and entanglement, QML models embed classical data into high-dimensional Hilbert spaces that may enable more efficient representation of nonlinear decision boundaries than classical feature mappings \cite{wittek2014quantum}. In the noisy intermediate-scale quantum (NISQ) era, variational quantum circuits (VQCs) and hybrid quantum–classical architectures have been the primary paradigms examined \cite{benedetti2019parameterized}. While VQCs are theoretically expressive, their practical performance is limited by optimization challenges such as barren plateaus, particularly for deeper circuits \cite{zhang2022fundamental}.
Hybrid quantum–classical approaches address these challenges by using quantum circuits as nonlinear feature maps while delegating model training to classical algorithms \cite{havlivcek2019supervised}. This approach improves training stability and avoids direct optimization of large quantum parameter spaces, while still keeping the representational benefits of quantum embeddings \cite{perez2020data,schuld2021effect}.
Despite growing interest, quantum machine learning for power-system cybersecurity has not been widely studied, especially in realistic attack scenarios \cite{eskandarpour2019quantum}. Most existing studies rely on simplified datasets and do not examine coordinated stealth attacks on distributed generation units operating in microgrids \cite{dunjko2018machine,ciliberto2018quantum}. This study investigates the application of quantum machine learning to detect coordinated stealth attacks targeting a distributed generation unit. Using high-fidelity simulated measurements of voltage magnitude, reactive power, and frequency deviation, supervised binary classifiers are trained to distinguish normal operation from malicious control disturbances \cite{wang2013cyber}. Variational quantum classifiers, hybrid quantum–classical models, and strong classical baselines are evaluated and compared \cite{sharma2022trainability,spall2002multivariate}. The results provide insight into the current capabilities and limitations of quantum learning methods and highlight their potential role in future cyber-resilient power-system architectures \cite{montanaro2016quantum,ajagekar2019quantum}.

The remainder of this paper is organized as follows: Section II reviews related work on cyber–physical attack detection in power systems and quantum machine learning. Section III describes the methodology, including the distributed generation model, coordinated stealth attack formulation, dataset construction, and the classical, quantum, and hybrid learning approaches. Section IV presents the experimental results and performance evaluation. Section V discusses the findings and practical implications for power-system cybersecurity. Section VI concludes the paper.

\section{Related Works}

Research on cyber–physical security of power systems has established that coordinated attackers can manipulate control and measurement signals while remaining undetected by conventional residual-based detection mechanisms \cite{pasqualetti2013attack,sandberg2010security}. Pasqualetti et al. characterized stealth attack construction using estimator null-space properties, providing a formal framework for undetectable attack design \cite{pasqualetti2013attack}. Sandberg et al. proposed vulnerability measures that show how exposed a system is to coordinated attacks. These measures make it possible to evaluate attack risks in networked control systems \cite{sandberg2010security}. Later studies applied these ideas to distributed generation and microgrids and showed that local controllers and communication links increase the number of possible attack points \cite{sridhar2011cyber,teixeira2015secure}.

Data-driven intrusion detection has been widely explored as a countermeasure to such attacks. Classical machine-learning methods, such as logistic regression, support vector machines, decision trees, and ensemble models, have shown strong performance in detecting cyberattacks and abnormal behavior in power systems \cite{he2016cyber,ozay2015machine}. Kernel-based SVMs have been particularly effective for nonlinear classification; however, prior work reports degraded sensitivity when attack signals are deliberately constrained to lie near normal operating manifolds, a defining characteristic of coordinated stealth attacks \cite{ozay2015machine}.

Recent advances in quantum machine learning have introduced alternative representations for nonlinear classification through quantum feature embeddings. Havlíček et al. proposed quantum-enhanced feature spaces capable of implicitly representing complex correlations beyond classical kernels \cite{havlivcek2019supervised}, while Schuld and Killoran established the theoretical relationship between quantum embeddings and kernel methods \cite{schuld2019quantum}. Variational quantum classifiers were subsequently investigated as trainable quantum models for supervised learning \cite{farhi2018classification,schuld2020circuit}. Despite their expressiveness, multiple studies have identified barren plateau phenomena and optimization instability as key limitations in the noisy intermediate-scale quantum regime \cite{mcclean2018barren,cerezo2021cost}.

Hybrid quantum–classical learning architectures have been proposed to address these limitations by decoupling quantum feature extraction from classical optimization \cite{mitarai2018quantum,perez2020data}. In such approaches, quantum circuits are used exclusively as nonlinear feature maps, while classification is performed using classical models. Although hybrid methods have demonstrated improved training stability in benchmark learning tasks, their application to power-system cybersecurity remains limited. Existing studies often rely on simplified system models or synthetic datasets and do not evaluate performance under coordinated stealth attack scenarios targeting distributed generation units \cite{zhou2022noise,zhou2022quantum}. The present work addresses this gap by evaluating quantum and hybrid learning models using high-fidelity distributed generation measurements under realistic coordinated stealth attacks.

Table \ref{tab:related_work_taxonomy} summarizes a taxonomy of related work, highlighting the methodological focus and limitations of existing approaches relative to the present study.

\renewcommand{\arraystretch}{1.25}

\begin{table*}[!t]
\caption{Structured Overview of Prior Work on Cyberattack Detection and Quantum Learning in Power Systems}
\label{tab:related_work_taxonomy}
\centering
\footnotesize
\setlength{\tabcolsep}{4pt}
\renewcommand{\arraystretch}{1.25}
\begin{tabular}{p{3.4cm} p{4.2cm} p{4.4cm} p{4.2cm}}
\hline
\textbf{Category} &
\textbf{Representative Works} &
\textbf{Main Contributions} &
\textbf{Limitations} \\
\hline

Stealth attack modeling and vulnerability analysis &
Pasqualetti \textit{et al.}~\cite{pasqualetti2013attack},  
Sandberg \textit{et al.}~\cite{sandberg2010security} &
Formal construction of stealth attacks using estimator null spaces; vulnerability metrics for coordinated attacks &
Focus on detection limits rather than learning-based mitigation; not evaluated with data-driven classifiers \\

Cyber–physical security of DGs and microgrids &
Sridhar \textit{et al.}~\cite{sridhar2011cyber},  
Teixeira \textit{et al.}~\cite{teixeira2015secure} &
Analysis of cyberattack surfaces in distributed generation and microgrid control architectures &
Primarily analytical or control-theoretic; limited use of learning-based intrusion detection \\

Classical machine-learning-based intrusion detection &
He \textit{et al.}~\cite{he2016cyber},  
Ozay \textit{et al.}~\cite{ozay2015machine} &
Application of supervised learning methods (LR, SVM, DT, ensembles) for attack and anomaly detection &
Performance degrades for coordinated stealth attacks constrained near normal operating manifolds \\

Quantum feature embeddings and variational classifiers &
Havlíček \textit{et al.}~\cite{havlivcek2019supervised},  
Schuld and Killoran~\cite{schuld2019quantum},  
Farhi and Neven~\cite{farhi2018classification},  
Schuld \textit{et al.}~\cite{schuld2020circuit} &
Quantum-enhanced feature spaces and variational quantum classifiers for nonlinear classification &
Optimization instability and barren plateaus limit scalability on NISQ hardware \\

Hybrid quantum–classical learning models &
Mitarai \textit{et al.}~\cite{mitarai2018quantum},  
Pérez-Salinas \textit{et al.}~\cite{perez2020data} &
Decoupling quantum feature extraction from classical training improves stability &
Mostly evaluated on benchmark or synthetic datasets; not applied to power-system cyberattack detection \\

Quantum learning in power-system applications &
Zhou and Zhang~\cite{zhou2022noise},  
Zhou \textit{et al.}~\cite{zhou2022quantum} &
Exploration of quantum computing and learning for power-system stability and analytics &
Did not consider coordinated stealth attacks or DG-focused intrusion detection \\

\hline
\end{tabular}
\end{table*}

\section{Methods}
This section explains the dataset construction process, the classical baseline models, the quantum data-encoding strategy, the variational quantum models, the hybrid quantum–classical feature-map architecture, and the optimization procedures used throughout the study. Mathematical formulations are included to give a full description of the quantum learning methods applied to the detection task.

\subsection{Distributed Generation Measurement Model}
\begin{figure}[ht]
    \centering
    \includegraphics[width=\linewidth]{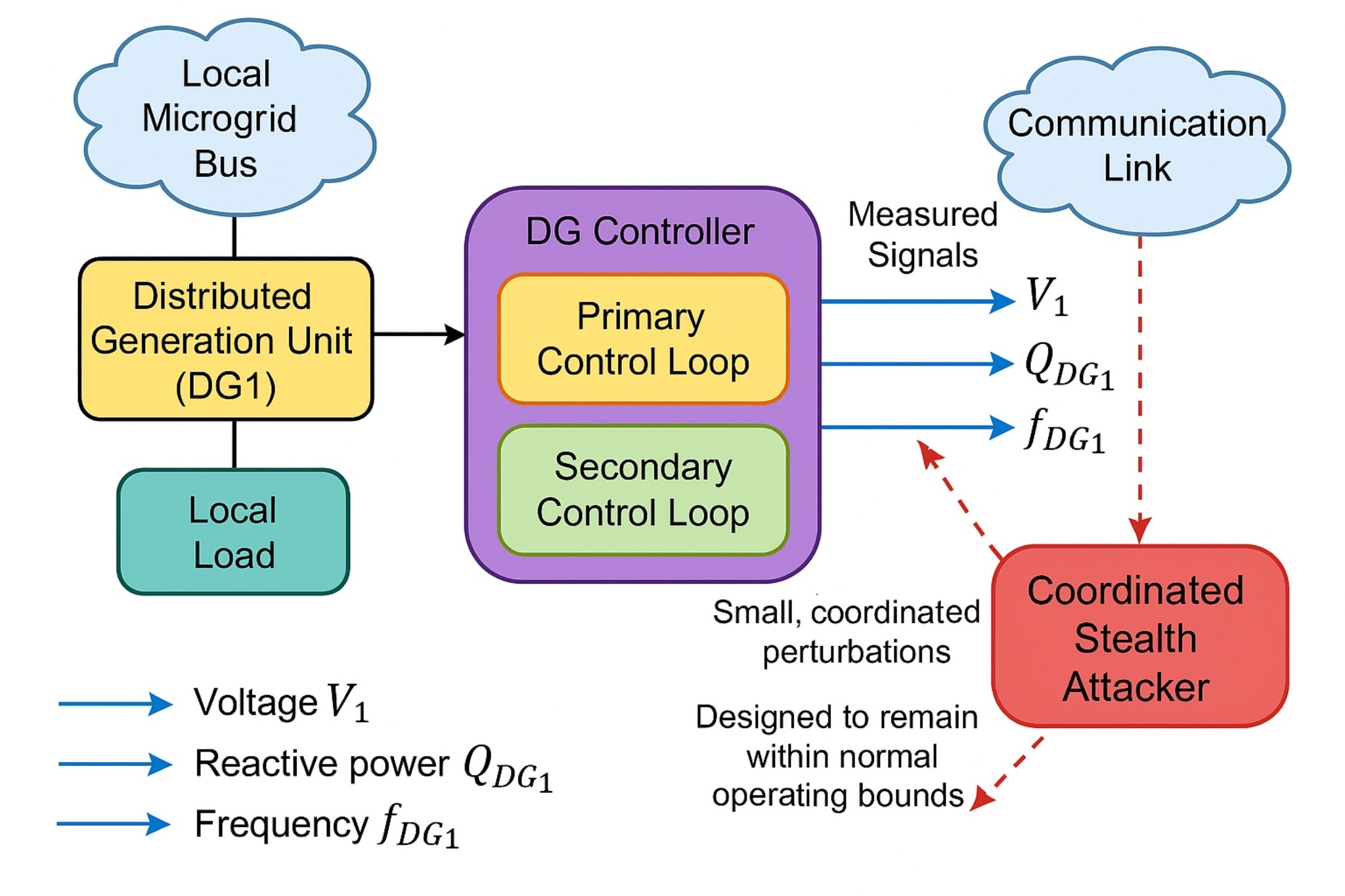}
    \caption{Distributed generation system and coordinated stealth attack model. An attacker injects small, coordinated perturbations into voltage magnitude, reactive power, and frequency measurements through compromised communication links while remaining within normal operating bounds to evade residual-based detection.}
    \label{fig:system_model}
\end{figure}

The distributed generation unit in this study operates under a hierarchical control architecture with primary and secondary control loops, as shown in Fig. \ref{fig:system_model}. Voltage magnitude, reactive power, and frequency measurements are exchanged over communication links and are vulnerable to cyber manipulation.
At each sampling instant \textit{$t_k$}, the DG1 subsystem reports its terminal voltage, reactive power, and frequency measurements. Let \textit{$V_1$} be the terminal voltage magnitude, $Q_{\mathrm{DG1}}$ (\textit{$t_k$}) be the reactive power injection, and $f_{\mathrm{DG1}}$ (\textit{$t_k$}) be the DG frequency measurement \cite{phadke2008synchronized,wang2013cyber}. These quantities constitute the raw measurement vector:

\begin{equation}
z(t_k) =
\begin{bmatrix}
V_1(t_k) \\
Q_{\mathrm{DG}1}(t_k) \\
f_{\mathrm{DG}1}(t_k)
\end{bmatrix}
\end{equation}
A frequency deviation feature is computed as
\begin{equation}
\Delta f(t_k) = f_{\mathrm{DG}1}(t_k) - f_0 ,
\end{equation}
where \( f_0 = 50~\mathrm{Hz} \) is the nominal frequency. The simulated microgrid operates at a nominal frequency of 50~Hz, consistent with the base frequency used in the MATLAB/Simulink model.

To enrich temporal information, first-order differences were computed:
\begin{equation}
\Delta z(t_k) = z(t_k) - z(t_{k-1}) .
\end{equation}

\begin{equation}
\Delta Q(t_k) = Q_{\mathrm{DG}1}(t_k) - Q_{\mathrm{DG}1}(t_{k-1})
\end{equation}

\begin{equation}
\Delta V(t_k) = V_1(t_k) - V_1(t_{k-1})
\end{equation}

Exploratory analysis was performed on candidate measurements. Active power was not retained in the final models, while reactive power, frequency deviation, and voltage magnitude were retained to align with the final dataset construction used in the learning pipeline. The final feature vector used for all classical, quantum, and hybrid models consists of reactive power and frequency-based measurements only. The final feature vector used in all experiments is defined as
\begin{equation}
\mathbf{x}(t_k) =
\begin{bmatrix}
Q_{\mathrm{DG}1}(t_k) \\
f_{\mathrm{dev}}(t_k) \\
V_1(t_k)
\end{bmatrix},
\end{equation}
where
\begin{equation}
f_{\mathrm{dev}}(t_k) = f_{\mathrm{DG}1}(t_k) - f_0,
\end{equation}
and $f_0 = 50~\mathrm{Hz}$ is the nominal frequency of the simulated microgrid.

\subsection{Coordinated Stealth Attack Model}
\begin{figure}[ht]
    \centering
    \includegraphics[width=\linewidth]{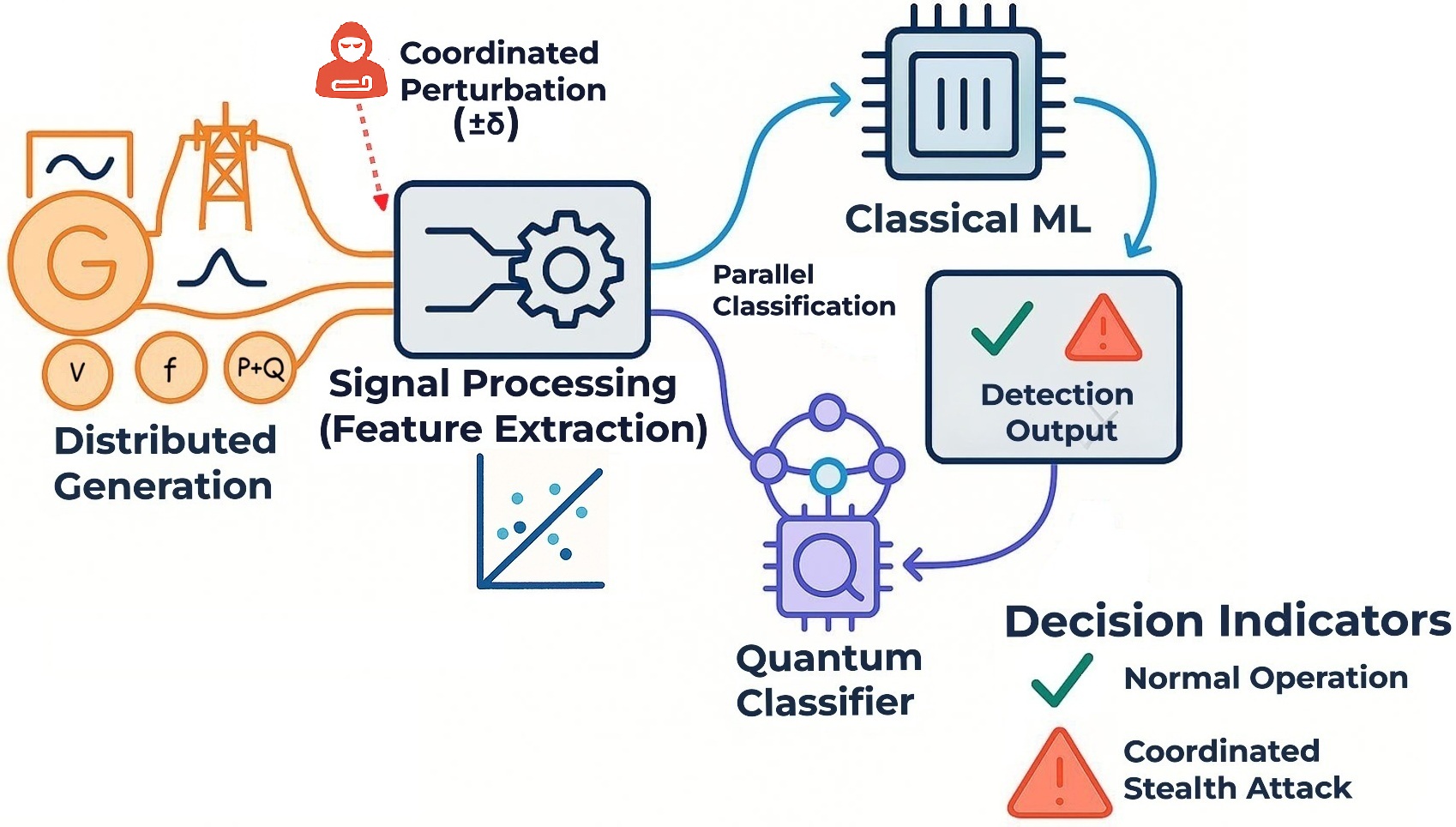}
    \caption{Parallel classical and quantum intrusion detection architecture for coordinated stealth attack detection in a distributed generation system. Voltage magnitude, frequency, and power measurements are processed to extract features, while small coordinated perturbations remain within normal operating bounds. Classical machine-learning models and a quantum classifier operate in parallel, producing a binary detection decision indicating normal operation or coordinated stealth attack.}
    \label{fig:Parallel_classification}
\end{figure}

Figure \ref{fig:Parallel_classification} illustrates the overall intrusion detection architecture, showing how coordinated stealth perturbations propagate through feature extraction and parallel classical and quantum classifiers to produce the final detection decision. Coordinated stealth attackers add small, carefully chosen changes to the signals while keeping them within normal operating limits, as previously shown in Fig \ref{fig:system_model}. This allows the attack to avoid detection by traditional residual-based methods.

Coordinated stealth attacks are designed to manipulate control behavior while preserving measurement patterns that closely resemble normal operating conditions. In this study, stealth attacks are implemented at the distributed secondary control layer rather than at the sensor level, following the cyber--physical modeling approach used in prior virtual microgrid studies \cite{liu2011false,srivastava2020microgrid,rath2020cyber}.

Let the uncompromised DG1 measurement vector at sampling instant $t_k$ be
\begin{equation}
z(t_k) =
\begin{bmatrix}
V_1(t_k) \\
Q_{\mathrm{DG}1}(t_k) \\
f_{\mathrm{DG}1}(t_k)
\end{bmatrix}.
\end{equation}

During a coordinated stealth attack, small, correlated perturbations are injected into the secondary control correction signals that regulate frequency and voltage restoration. These perturbations propagate through the hierarchical control loops and manifest as subtle deviations in voltage, reactive power, and frequency measurements. The injected perturbation vector is expressed as
\begin{equation}
a(t_k) =
\begin{bmatrix}
a_V(t_k) \\
a_Q(t_k) \\
a_f(t_k)
\end{bmatrix},
\end{equation}
resulting in corrupted measurements
\begin{equation}
\tilde{z}(t_k) = z(t_k) + a(t_k).
\end{equation}

Rather than explicitly solving a measurement Jacobian null-space condition, stealthiness is achieved by constraining the injected perturbations to remain within normal operating limits and by preserving correlations among control variables. This ensures that conventional residual-based or threshold-based detection mechanisms are unable to reliably distinguish the attack from normal system behavior. Similar physically consistent stealth strategies have been shown to evade traditional detection methods in distributed microgrid control architectures \cite{liu2011false,rath2020cyber}.

Each sample is assigned a binary label
\begin{equation}
y(t_k) \in \{0,1\},
\end{equation}
where $y=0$ denotes normal operation and $y=1$ denotes a coordinated stealth attack.

\begin{figure}[ht]
\centering
\includegraphics[width=\linewidth]{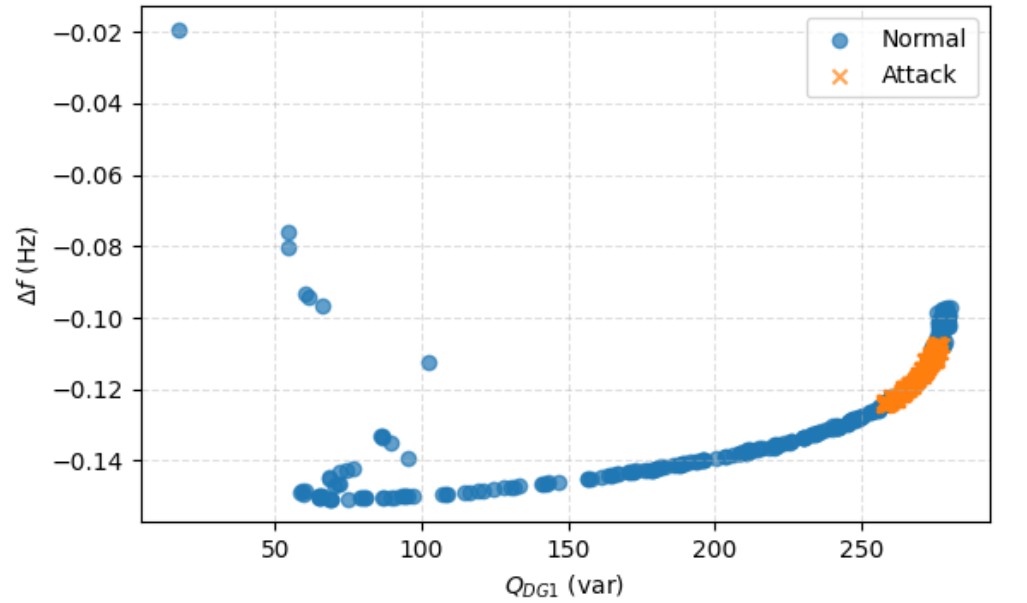}
\caption{DG1 Measurements under Normal Operation and Coordinated Stealth Attack.}
\label{fig:DG1_measurements}
\end{figure}

Fig.~\ref{fig:DG1_measurements} shows the relationship between reactive power at DG1 ($Q_{\mathrm{DG1}}$) and frequency deviation ($\Delta f$) under normal and coordinated stealth attack conditions. Both cases follow a similar nonlinear pattern, confirming that the attack closely resembles normal behavior. This overlap explains why stealth attacks are difficult to detect and why simple threshold-based methods do not work well. At the same time, the consistent shift in the attack data provides a useful structure that advanced machine-learning and quantum-enhanced models can learn to exploit.

\subsection{Feature Normalization and Quantum Scaling}
To ensure numerical stability and fair model evaluation, feature normalization was applied using training data statistics only. Let $x_i(t_k)$ denote the $i$th raw feature at time $t_k$. Each feature was first standardized using z-score normalization,
\begin{equation}
x_i^{\mathrm{norm}}(t_k) =
\frac{x_i(t_k) - \mu_i}{\sigma_i},
\end{equation}
where $\mu_i$ and $\sigma_i$ are the mean and standard deviation of feature $x_i$, computed exclusively from the training set.

For quantum angle encoding, the normalized features were further rescaled to valid rotation angles using min--max scaling,
\begin{equation}
x_i^{q}(t_k) =
\left(
\frac{x_i^{\mathrm{norm}}(t_k) - x_{i,\min}}
{x_{i,\max} - x_{i,\min}}
\right)\pi
- \frac{\pi}{2},
\end{equation}
where $x_{i,\min}$ and $x_{i,\max}$ are the minimum and maximum values of the normalized feature $x_i^{\mathrm{norm}}$, again computed from the training set only.

This two-stage normalization prevents data leakage from the test set, ensures numerical stability for classical baseline models, and maps features to valid quantum rotation parameters required for angle encoding \cite{schuld2019quantum}. All normalization parameters were fixed after training-set estimation and reused unchanged during validation and testing.

\subsection{Quantum Data Encoding}

An angle encoding was employed, which is also called rotation encoding. Let the classical feature vector be
\begin{equation}
\mathbf{x} = [x_1, x_2, x_3]^{\top}.
\end{equation}.

In this study, $(x_1,x_2,x_3)$ correspond to $(Q_{\mathrm{DG}1},\, f_{\mathrm{dev}},\, V_1)$ after scaling to valid rotation angles.

This vector is encoded into a 3-qubit quantum state using the data encoding unitary operator:

\begin{equation}
U_{\mathrm{enc}}(\mathbf{x}) = \prod_{i=1}^{3} R_y(x_i)^{(i)},
\end{equation}
where $R_y(x_i)$ denotes a single-qubit rotation about the $Y$-axis applied on qubit $i$ with angle $x_i$.

The resulting encoded quantum state is given by: 
\begin{equation}
\lvert \psi(\mathbf{x}) \rangle = U_{\mathrm{enc}}(\mathbf{x}) \lvert 0 \rangle^{\otimes 3},
\end{equation}
Where  \( \lvert 0 \rangle^{\otimes 3} \) is the three-qubit computational ground state. An entangling unitary operator is introduced using \begin{equation}
U_{\mathrm{ent}} = \mathrm{CNOT}_{1 \rightarrow 2}\,\mathrm{CNOT}_{2 \rightarrow 3},
\end{equation}
where \(\mathrm{CNOT}_{i \rightarrow j}\) is a controlled-NOT gate with control qubit \( i \) and target qubit \( j \).
Thus, the encoded quantum state is given by
\begin{equation}
\lvert \phi(\mathbf{x}) \rangle
= U_{\mathrm{ent}}\,U_{\mathrm{enc}}(\mathbf{x}) \lvert 0 \rangle^{\otimes 3}.
\end{equation} 
Angle encoding combined with entanglement enables nonlinear feature representations in the quantum Hilbert space \cite{perez2020data}.

\subsection{Quantum Circuit Architecture}
All quantum models were implemented using a three-qubit circuit, corresponding to the three-dimensional classical feature vector
\(
[Q_{\mathrm{DG}1},\; f_{\mathrm{dev}},\; V_1]
\).
Each feature was encoded using single-qubit $R_y$ rotation gates, followed by a parameterized variational block. The VQC consists of $L$ repeated layers. Each layer applies parameterized single-qubit rotations followed by a ladder-style entangling operation. Specifically, the $l$th layer is defined as

\begin{equation}
\begin{aligned}
U_l(\boldsymbol{\theta}_l)
&=
\Bigg(
\bigotimes_{i=1}^{3} R_y(\theta_{l,i})
\Bigg)
\, \mathrm{CNOT}_{1 \rightarrow 2}\,
\mathrm{CNOT}_{2 \rightarrow 3}
\end{aligned}
\end{equation}

\begin{equation}
\bigotimes_{i=1}^{3} R_y(\theta_{l,i})
=
R_y(\theta_{l,1}) \otimes R_y(\theta_{l,2}) \otimes R_y(\theta_{l,3}),
\end{equation}

where $\boldsymbol{\theta}_l = [\theta_{l,1}, \theta_{l,2}, \theta_{l,3}]$ represents the trainable parameters of layer $l$.

The total number of trainable parameters is $3L$. Shallow ($L=1$), medium ($L=2$), and deep ($L=3$) circuits were evaluated to study the effect of circuit depth on classification performance.

The circuit output was obtained by measuring the Pauli-$Z$ observable on the first qubit, i.e., $\hat{O} = Z \otimes I \otimes I$. Quantum experiments were executed using Qiskit software-based simulators that compute expectation values without access to dedicated quantum hardware. Hybrid quantum features were obtained using exact statevector simulation to compute Pauli--$Z$ expectation values and multi-qubit correlations, while VQC models were trained using Qiskit's \texttt{EstimatorQNN} framework.

Table~\ref{tab:quantum_config} summarizes the quantum circuit structure and feature map configuration used for the variational and hybrid quantum models. This table gives architectural details that help with reproducibility and explain how classical measurements are mapped into quantum representations.

\begin{table}[!t]
\caption{Quantum Circuit and Feature Map Configuration}
\label{tab:quantum_config}
\centering
\footnotesize
\setlength{\tabcolsep}{4pt}
\renewcommand{\arraystretch}{1.2}
\begin{tabular}{p{3.2cm} p{3.8cm}}
\hline
\textbf{Component} & \textbf{Configuration} \\
\hline
Classical feature vector 
& $\left[ Q_{\mathrm{DG1}},\; f_{\mathrm{dev}},\; V_1 \right]$ \\

Number of qubits 
& 3 (one qubit per feature) \\

Data encoding 
& Angle encoding using $R_y$ rotations \\

Entanglement structure 
& Ladder CNOT gates ($\mathrm{CNOT}_{1 \rightarrow 2}$, $\mathrm{CNOT}_{2 \rightarrow 3}$) \\

Variational circuit depth 
& $L = 1,\; 2,\; 3$ \\

Trainable parameters 
& $3L$ \\

Measurement observable (VQC) 
& $Z \otimes I \otimes I$ \\

Hybrid quantum feature extraction 
& \begin{tabular}{@{}l@{}}
$\langle ZII\rangle,\; \langle IZI\rangle,\; \langle IIZ\rangle,\langle ZZI\rangle,\;$ \\
$\langle IZZ\rangle,\; \langle ZIZ\rangle,\; \langle ZZZ\rangle$ (7D)
\end{tabular} \\

Quantum backend 
& Qiskit simulator (no quantum hardware) \\
\hline
\end{tabular}
\end{table}

\subsection{Variational Quantum Classifier (VQC)}
The VQC consists of a parameterized quantum circuit composed of single-qubit rotation gates and entangling operations. The trainable unitary is expressed as
\begin{equation}
U(\boldsymbol{\theta}) = \prod_{\ell=1}^{L} U_{\ell}(\boldsymbol{\theta}_{\ell}),
\end{equation}
where $\boldsymbol{\theta}$ denotes the set of trainable parameters and $L$ is the circuit depth.

The circuit output is obtained by measuring the Pauli-$Z$ observable on the first qubit. For the two-qubit circuit used in this study, the measurement operator is defined as
\begin{equation}
\hat{O} = Z \otimes I .
\end{equation}

The VQC produces a continuous decision score given by the expectation value
\begin{equation}
s(\mathbf{x}) =
\langle \phi(\mathbf{x}) \rvert
U^{\dagger}(\boldsymbol{\theta}) \,
\hat{O} \,
U(\boldsymbol{\theta})
\lvert \phi(\mathbf{x}) \rangle,
\end{equation}
where $s(\mathbf{x}) \in [-1,1]$.

Binary class labels are obtained by thresholding the continuous score,
\begin{equation}
\hat{y} =
\begin{cases}
1, & s(\mathbf{x}) \ge 0, \\
0, & s(\mathbf{x}) < 0.
\end{cases}
\end{equation}

Model parameters are trained by minimizing the mean squared error (MSE) loss,
\begin{equation}
\mathcal{L}(\boldsymbol{\theta}) =
\frac{1}{N}
\sum_{k=1}^{N}
\left( y_k - s(\mathbf{x}_k) \right)^2 ,
\end{equation}
where $N$ is the number of training samples and $y_k \in \{0,1\}$ denotes the true class label. The MSE loss is commonly used in variational quantum classifiers because it directly penalizes deviations between the continuous expectation-value output and the target labels while remaining stable under noisy gradient estimates in NISQ-era optimization \cite{benedetti2019parameterized,farhi2018classification}.

\subsection{Optimization Algorithms}
Two gradient-free optimization methods were used to train the variational quantum classifier: Constrained Optimization BY Linear Approximations (COBYLA) and simultaneous perturbation stochastic approximation (SPSA). Gradient-free methods are preferred in NISQ-era quantum learning due to noisy objective evaluations and the absence of reliable analytic gradients.

COBYLA performs constrained optimization using local linear approximations of the loss function and requires only function evaluations. It was used primarily for shallow circuits due to its fast initial convergence. However, its performance degraded as circuit depth increased.

To improve robustness for deeper circuits, SPSA was employed. SPSA estimates the gradient using only two stochastic loss evaluations per iteration,
\begin{equation}
\hat{g}_{k,i} =
\frac{
\mathcal{L}\!\left(\boldsymbol{\theta}_k + c_k \boldsymbol{\Delta}_k\right)
-
\mathcal{L}\!\left(\boldsymbol{\theta}_k - c_k \boldsymbol{\Delta}_k\right)
}{
2 c_k \Delta_{k,i}
},
\end{equation}
where $\boldsymbol{\Delta}_k$ is a random perturbation vector with independent symmetric Bernoulli entries. Model parameters are updated according to
\begin{equation}
\boldsymbol{\theta}_{k+1}
=
\boldsymbol{\theta}_{k}
-
a_k \hat{\boldsymbol{g}}_k .
\end{equation}

In this study, COBYLA was run for up to 80 iterations for shallow circuits, while SPSA was used for medium and deep circuits with a maximum of 200 iterations. The SPSA gain sequences $a_k$ and $c_k$ were selected using standard diminishing step-size schedules to balance exploration and convergence stability \cite{spall2002multivariate}.

\subsection{Hybrid Quantum--Classical Feature Map}

In the hybrid approach, the quantum circuit is used solely as a nonlinear feature map rather than a trainable end-to-end classifier. 
First, expectation values of local Pauli-$Z$ operators are extracted as base quantum features:
\begin{equation}
z_j(\mathbf{x}) =
\langle \phi(\mathbf{x}) \rvert \hat{O}_j \lvert \phi(\mathbf{x}) \rangle ,
\end{equation}
where the local measurement operators are defined as
\begin{equation}
\hat{O}_1 = Z \otimes I \otimes I, \quad
\hat{O}_2 = I \otimes Z \otimes I, \quad
\hat{O}_3 = I \otimes I \otimes Z .
\end{equation}

These three Pauli expectation values form the base quantum features
\(
(z_1, z_2, z_3)
\).
To improve the capacity of the classical classifiers, the final hybrid pipeline includes simple nonlinear interaction terms in the feature set. 
The resulting hybrid feature vector is seven-dimensional and defined as
\begin{equation}
\mathbf{z}_q(\mathbf{x}) =
\begin{bmatrix}
z_1(\mathbf{x})\\
z_2(\mathbf{x})\\
z_3(\mathbf{x})\\
z_1(\mathbf{x})z_2(\mathbf{x})\\
z_1(\mathbf{x})z_3(\mathbf{x})\\
z_2(\mathbf{x})z_3(\mathbf{x})\\
z_1(\mathbf{x})z_2(\mathbf{x})z_3(\mathbf{x})
\end{bmatrix}.\end{equation}

This augmented representation keeps the original quantum observables interpretable, while giving classical models access to higher-order relationships produced by the quantum feature map. These hybrid features are then used to train classical classifiers such as logistic regression and support vector machines:
\begin{equation}
\hat{y} = f_{\mathrm{classical}}\!\left( \mathbf{z}_q(\mathbf{x}) \right).\end{equation}

Hybrid quantum–classical models improve training stability by avoiding direct optimization of large quantum parameter spaces, while still keeping the nonlinear benefits of quantum encoding \cite{schuld2021effect}. All quantum circuits and learning models were implemented using Qiskit \cite{qiskit2019zenodo}, with circuit operations and measurements defined using OpenQASM \cite{cross2017open}.

\subsection{Evaluation Metrics}
Binary classification performance is evaluated using:
\begin{equation}
\mathrm{Accuracy} =
\frac{TP + TN}{TP + TN + FP + FN}.
\end{equation}

\begin{equation}
\mathrm{F1\text{-}Score} =
\frac{2TP}{2TP + FP + FN}.
\end{equation}

The confusion matrix is expressed as:
\begin{equation}
\mathrm{Confusion\ Matrix} =
\begin{bmatrix}
TN & FP \\
FN & TP
\end{bmatrix}.
\end{equation}
where TN represents true negatives, FP represents false positives, FN represents false negatives, and TP represents true positives \cite{fawcett2006introduction,bishop2006pattern}. FNs are important in cyberattack detection because they represent attacks that occur but are not detected by the model.

\subsection{Computational Environment}

The microgrid model and coordinated stealth attack datasets were developed using MATLAB/Simulink on a local workstation running Windows 11 with a 13th-generation Intel Core i9-13900HX processor and 16 GB of RAM. This environment was used exclusively for power system modeling and dataset generation. All classical and quantum machine-learning experiments were conducted on a virtualized x86\_64 computing environment provided through Google Colab, running on an Intel Xeon processor operating at 2.20~GHz with two logical CPU cores and KVM-based hardware virtualization. Quantum experiments were executed using software-based quantum backends, without access to dedicated quantum hardware.

\subsection{Evaluation Protocol and Reproducibility}
All experiments used a fixed 70\% / 30\% train--test split with a random seed of 42 to ensure reproducible results. Feature normalization parameters were computed using the training data only and applied unchanged to the test data. Classical, quantum-only, and hybrid models were evaluated using the same training and testing sets to allow fair comparison. Because variational quantum training is stochastic, different runs can produce slightly different results. The reported performance metrics correspond to representative runs that showed stable convergence, as reflected in the training loss curves. This evaluation setup ensures that performance differences are due to the learning models and feature representations, rather than differences in data splitting.

\section{Result}
This section presents the performance of classical machine learning baselines, quantum machine learning models, and hybrid quantum–classical approaches on the coordinated stealth attack dataset. The evaluation uses the accuracy, F1 score, and confusion matrix metrics.

\subsection{Dataset Summary}
A balanced dataset of 600 samples was constructed, consisting of 300 normal operating points and 300 coordinated stealth attack samples. Each sample is represented by three physically meaningful features: reactive power $Q_{\mathrm{DG}1}$, frequency deviation $f_{\mathrm{dev}}$, and voltage magnitude $V_1$. These features capture subtle control-loop disturbances introduced by coordinated stealth attacks while remaining within normal operating bounds. The dataset was divided into 420 training samples and 180 test samples using a stratified 70\%/30\% split.

\begin{figure}[ht]
    \centering
    \includegraphics[width=\columnwidth]{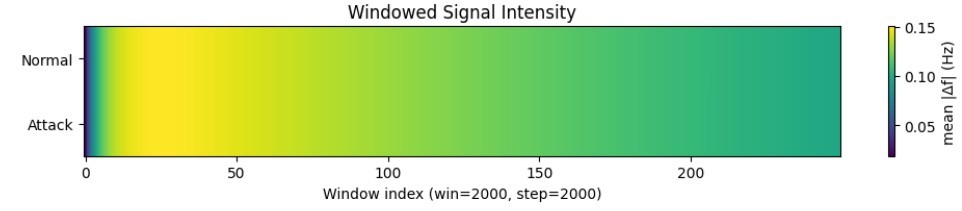}
    \caption{Windowed analysis of frequency deviation magnitude using non-overlapping sample windows ($\text{win}=2000$, $\text{step}=2000$). Each row represents the mean absolute frequency deviation for normal operation and coordinated stealth attack conditions. The similarity across windows highlights the stealthy temporal behavior of the attack and motivates the use of feature-based learning methods.}
    \label{fig:windowed_heatmap}
\end{figure}

Figure~\ref{fig:windowed_heatmap} examines the temporal behavior of coordinated stealth attacks using a windowed aggregation of frequency deviation magnitude. The close similarity between normal and attack sequences across all windows shows that the injected perturbations remain within nominal operating bounds over time, reinforcing the difficulty of detection using threshold-based or residual-based methods.

\subsection{Classical Baseline Performance}

\begin{figure}[t]
    \centering
    \begin{minipage}{0.48\columnwidth}
        \centering
        \includegraphics[width=\linewidth]{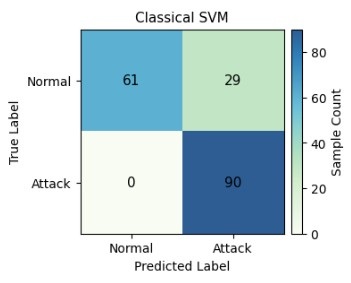}\\
        {\footnotesize (a)}
    \end{minipage}
    \hfill
    \begin{minipage}{0.48\columnwidth}
        \centering
        \includegraphics[width=\linewidth]{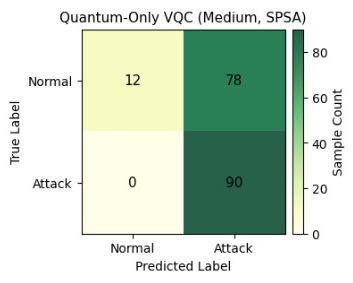}\\
        {\footnotesize (b)}
    \end{minipage}

    \vspace{6pt}

    \begin{minipage}{0.6\columnwidth}
        \centering
        \includegraphics[width=\linewidth]{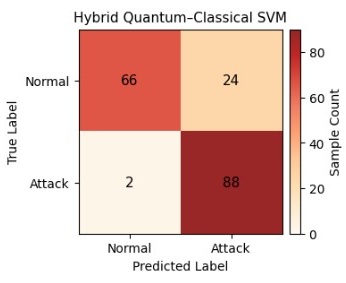}\\
        {\footnotesize (c)}
    \end{minipage}

    \caption{%
    Confusion matrices for intrusion detection models:
    (a) Classical SVM,
    (b) variational quantum classifier,
    and (c) hybrid quantum--classical SVM.
    }
    \label{fig:Confusion_matrices}
\end{figure}

Classical machine-learning models were trained using normalized DG measurements as a reference for comparison. Logistic regression reached an accuracy of 0.761 and an F1 score of 0.807, showing stable but limited class separation. Table \ref{tab:performance_comparison} shows that the RBF-kernel support vector machine achieved the best classical performance, with an accuracy of 0.839 and an F1 score of 0.861. As shown in Fig.~\ref{fig:Confusion_matrices}(a), the classical SVM correctly detected all attack instances, forming a strong and stable baseline for evaluating quantum-only and hybrid learning methods.

\subsection{Quantum-Only Variational Classifiers}
The VQCs were tested using a three-qubit angle-encoding scheme with ladder-style entanglement, different circuit depths, and both COBYLA and SPSA optimizers. The shallow VQC could not learn a useful decision boundary and performed close to random guessing, with an accuracy of 0.500. Increasing the circuit depth led to only small improvements. As shown in Fig.~\ref{fig:Confusion_matrices}(b), the medium-depth VQC trained with SPSA gave the best performance among the quantum-only models, achieving an accuracy of 0.606 and an F1 score of 0.717, as reported in Table \ref{tab:performance_comparison}. Using deeper circuits reduced performance, which is consistent with training instability and barren plateau effects.

\begin{figure}[ht]
    \centering
    \includegraphics[width=\linewidth]{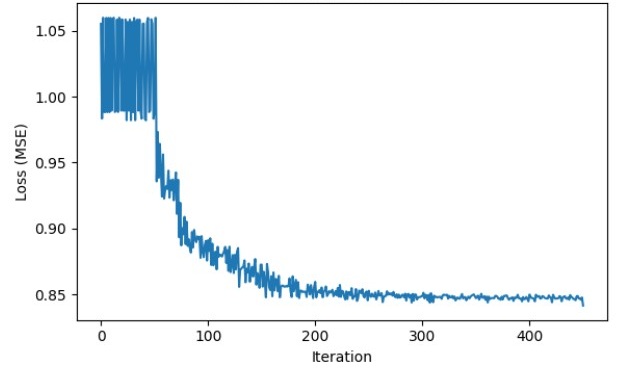}
    \caption{Training loss of the VQC optimized using the SPSA algorithm. The loss exhibits an initial high-variance exploration phase, followed by rapid convergence and stabilization at a local minimum, reflecting the stochastic nature of SPSA and the limited expressivity of shallow NISQ-era quantum circuits.}
    \label{fig:Training_loss}
\end{figure}

Figure \ref{fig:Training_loss} shows the training behavior of the VQC using the SPSA optimizer. At the beginning of training, the loss changes widely because SPSA relies on random parameter updates and noisy gradient estimates. After about 50–100 iterations, the optimizer finds a good direction, and the loss drops quickly. After this, the loss stays near 0.84, suggesting that the model has converged to a local minimum.

This flat region indicates that learning is limited by the low expressivity of the shallow quantum circuit and the noisy optimization process typical of NISQ-era devices. Similar behavior has been observed in other variational quantum algorithms, where increasing circuit depth often causes unstable training or barren plateaus. Overall, the quantum-only VQC is stable but performs worse than classical models, which motivates the use of hybrid quantum–classical methods that combine quantum feature extraction with classical classifiers.

\begin{table}[!t]
\caption{Performance Comparison of Classical, Quantum-Only, and Hybrid Models}
\label{tab:performance_comparison}
\centering
\footnotesize
\setlength{\tabcolsep}{3pt}
\begin{tabular}{>{\centering\arraybackslash}p{2.6cm} c c p{2.9cm}}
\hline
\textbf{Model} & \textbf{Accuracy} & \textbf{F1 Score} & \textbf{Observation} \\
\hline
Classical SVM (RBF) &
0.839 & 0.861 &
Strong baseline on low-dimensional features \\

Variational Quantum Classifier (SPSA) &
0.606 & 0.717 &
Learnable but limited by NISQ optimization \\

Hybrid Quantum--Classical SVM &
0.856 & 0.871 &
Best overall performance using quantum feature embedding \\
\hline
\end{tabular}
\end{table}

\subsection{Hybrid Quantum–Classical Feature Models}

To overcome the optimization limits of fully variational quantum classifiers, a hybrid quantum--classical approach was adopted in which quantum circuits act solely as nonlinear feature maps. Expectation values of Pauli-$Z$ operators and their multi-qubit correlations were extracted from a three-qubit quantum state, forming a seven-dimensional hybrid feature representation used for classical classification.

\subsubsection{Pairwise Quantum Feature Relationships}
\begin{figure}[ht]
    \centering
    \includegraphics[width=\linewidth]{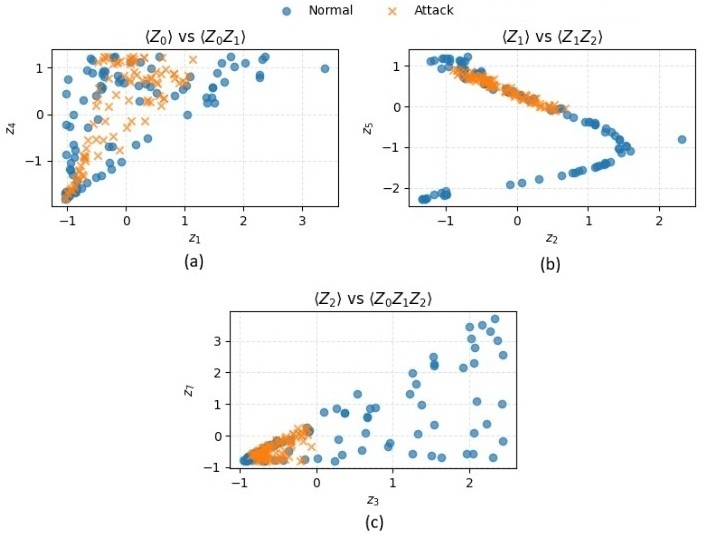}
    \caption{Pairwise relationships between hybrid quantum features extracted from a three-qubit quantum feature map. Panels (a)–(c) show scatter plots of single-qubit Pauli-$Z$ expectations versus higher-order correlations: (a) $\langle Z_0 \rangle$ vs.\ $\langle Z_0 Z_1 \rangle$, (b) $\langle Z_1 \rangle$ vs.\ $\langle Z_1 Z_2 \rangle$, and (c) $\langle Z_2 \rangle$ vs.\ $\langle Z_0 Z_1 Z_2 \rangle$. Normal and coordinated stealth attack samples form distinct nonlinear manifolds, indicating that entanglement-induced correlations improve class separability prior to classical classification.}
    \label{fig:Pairwise}
\end{figure}

\mbox{}\\
Figure~\ref{fig:Pairwise} shows the pairwise relationships between selected hybrid quantum features extracted from the three-qubit quantum feature map. Each panel shows the correlation between a single-qubit Pauli-$Z$ expectation value and a higher-order Pauli interaction term computed from the same quantum state.

The hybrid quantum features consist of single-qubit Pauli-$Z$ expectation values 
$z_1=\langle Z_0\rangle$, $z_2=\langle Z_1\rangle$, $z_3=\langle Z_2\rangle$, 
along with higher-order correlation terms 
$\langle Z_0Z_1\rangle$, $\langle Z_1Z_2\rangle$, and $\langle Z_0Z_1Z_2\rangle$
extracted from the same three-qubit quantum state.

In Fig.~\ref{fig:Pairwise}(a), the relationship between $\langle Z_0 \rangle$ and $\langle Z_0 Z_1 \rangle$ is shown. Both normal and coordinated stealth attack samples follow a curved nonlinear trend, indicating that entanglement introduces structured correlations between local and pairwise observables. Attack samples tend to occupy a more compact region along this manifold, while normal samples exhibit greater dispersion. Fig.~\ref{fig:Pairwise}(b) shows $\langle Z_1 \rangle$ versus $\langle Z_1 Z_2 \rangle$, where the data align along a narrow, nonlinear trajectory. This strong correlation comes from constraints imposed by the shared entanglement structure of the quantum circuit. Coordinated stealth attack samples are concentrated within a tighter segment of this trajectory, whereas normal samples extend over a wider range. In Fig.~\ref{fig:Pairwise}(c), the relationship between $\langle Z_2 \rangle$ and the three-body interaction term $\langle Z_0 Z_1 Z_2 \rangle$ is shown. Here, attack samples are clustered near low-magnitude values of the three-qubit correlation, while normal samples span a broader region of the feature space. This shows that higher-order quantum correlations capture small but consistent differences between normal and attack conditions.

\subsubsection{Marginal Distributions of Quantum Features}
\begin{figure}[ht]
    \centering
    \includegraphics[width=\linewidth]{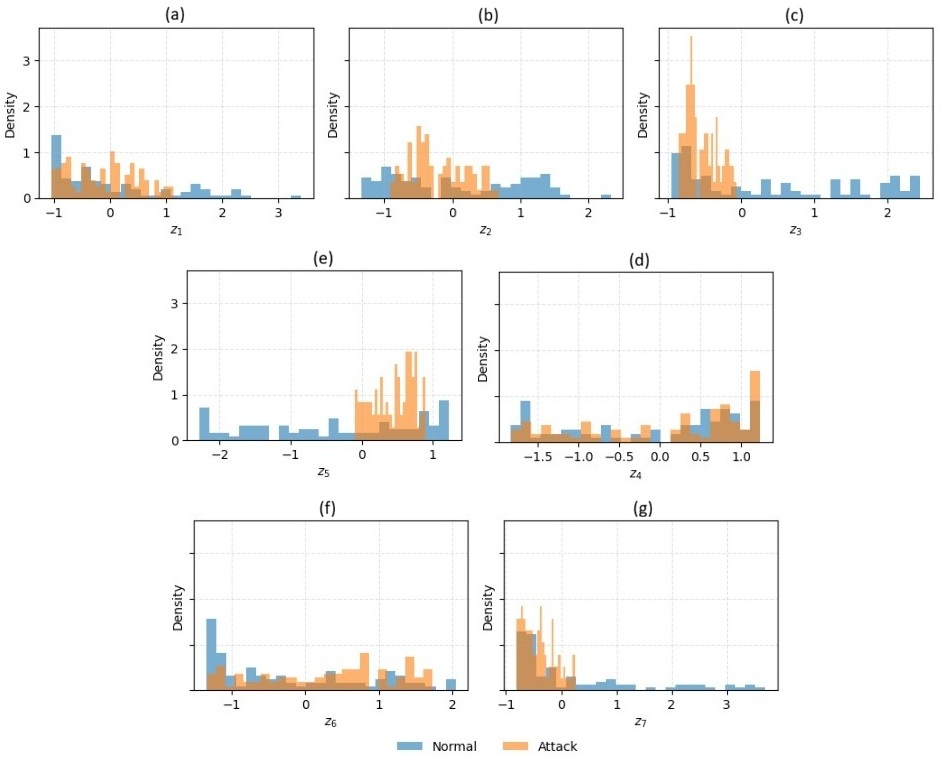}
    \caption{Marginal distributions of hybrid quantum features extracted from a three-qubit quantum feature map. Panels (a)–(c) show single-qubit Pauli-$Z$ expectation values $z_1=\langle Z_0\rangle$, $z_2=\langle Z_1\rangle$, and $z_3=\langle Z_2\rangle$. Panels (d)–(f) present two-qubit correlation terms $z_4=\langle Z_0Z_1\rangle$, $z_5=\langle Z_1Z_2\rangle$, and $z_6=\langle Z_0Z_2\rangle$, while panel (g) shows the three-qubit correlation $z_7=\langle Z_0Z_1Z_2\rangle$. Distributions are shown for normal operation and coordinated stealth attack conditions.}
    \label{fig:quantum_feature_distributions}
\end{figure}

\mbox{}\\
Figure~\ref{fig:quantum_feature_distributions} shows the marginal distributions of the seven hybrid quantum features obtained from the three-qubit quantum feature map. Panels (a)–(c) correspond to the single-qubit Pauli-$Z$ expectation values $z_1=\langle Z_0\rangle$, $z_2=\langle Z_1\rangle$, and $z_3=\langle Z_2\rangle$. Panels (d)–(f) show the two-qubit correlation terms $z_4=\langle Z_0 Z_1\rangle$, $z_5=\langle Z_1 Z_2\rangle$, and $z_6=\langle Z_0 Z_2\rangle$, while panel (g) illustrates the three-qubit correlation $z_7=\langle Z_0 Z_1 Z_2\rangle$.

Across multiple features, particularly $z_3$, $z_5$, and $z_7$, coordinated stealth attack samples cluster within narrow, high-density regions, whereas normal samples exhibit broader variability. This effect becomes more pronounced for higher-order correlation features, indicating that multi-qubit quantum measurements emphasize subtle dependencies between system variables not evident in the original measurement space.

\subsubsection{Comparison with Classical Feature Representations}
\begin{figure}[ht]
    \centering
    \begin{minipage}{0.48\columnwidth}
        \centering
        \includegraphics[width=\linewidth]{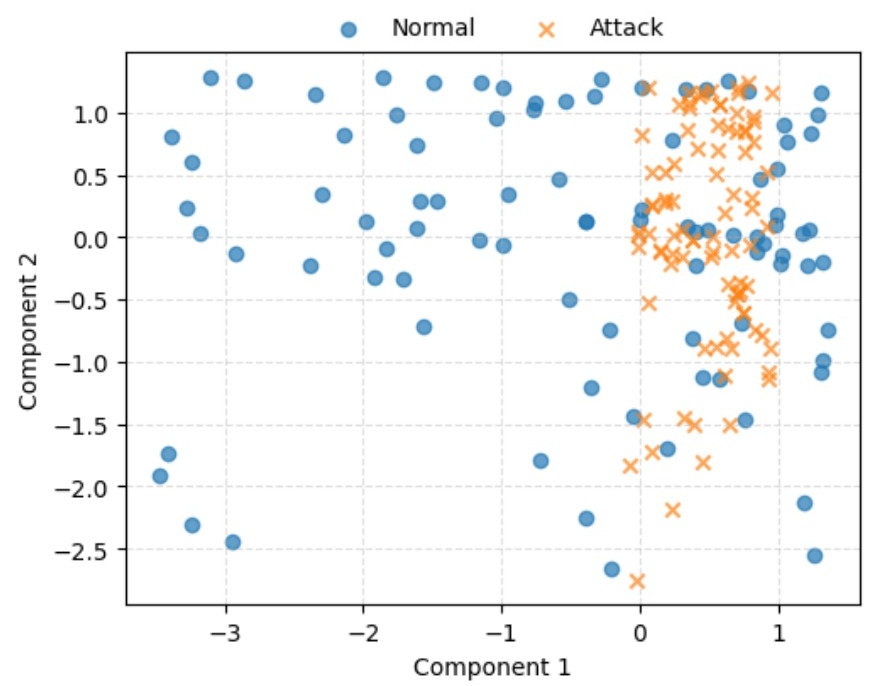}\\
        (a)
    \end{minipage}
    \hfill
    \begin{minipage}{0.48\columnwidth}
        \centering
        \includegraphics[width=\linewidth]{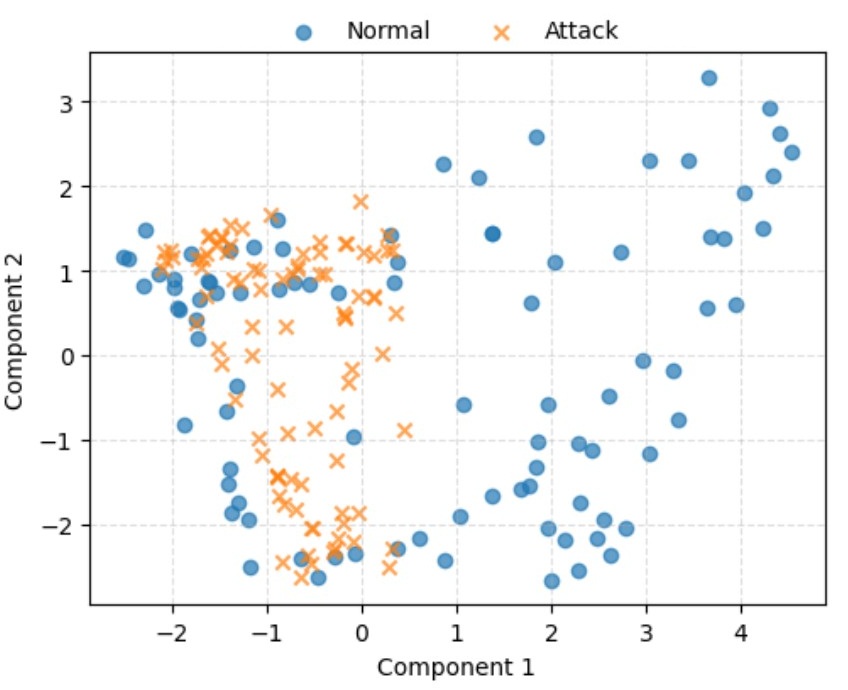}\\
        (b)
    \end{minipage}
    \caption{%
    Principal component analysis (PCA) of feature representations:
    (a) PCA of classical features $(Q_{\mathrm{DG}1}, \Delta f, V_1)$ and
    (b) PCA of hybrid quantum features obtained from Pauli-$Z$ expectation values and interaction terms.
    Normal operation samples and coordinated stealth attack samples are shown for comparison.
    }
    \label{fig:feature_embeddings}
\end{figure}

\mbox{}\\
Figures~\ref{fig:feature_embeddings}(a) and \ref{fig:feature_embeddings}(b) compare classical and hybrid quantum feature representations using PCA. In the classical feature space, normal and coordinated stealth attack samples largely overlap, showing that they are hard to separate even after dimensionality reduction.

In contrast, the PCA view of the hybrid quantum feature space shows clearer structure. Attack samples group together more closely, while normal samples spread out into different areas. This happens because the quantum feature map applies nonlinear transformations and interaction terms to the data. Although PCA is a linear method, the improved separation in the hybrid case suggests that quantum embeddings reshape the data in a way that makes it easier for classical classifiers to distinguish between normal and attack conditions, which supports the better detection performance of the hybrid quantum–classical models under NISQ constraints.

\subsubsection{Hybrid Classification Performance}
\mbox{}\\
Logistic regression trained on the augmented quantum feature set achieved an accuracy of 0.833 and an F1 score of 0.856, demonstrating that quantum feature embeddings improve linear separability relative to classical features. As shown in Table~\ref{tab:performance_comparison}, the hybrid quantum--classical SVM achieved the strongest overall performance, with an accuracy of 0.856 and an F1 score of 0.871, slightly outperforming the classical SVM baseline. Fig.~\ref{fig:Confusion_matrices}(c) shows that the hybrid model maintains high attack detection accuracy while reducing false positives compared to both quantum-only and classical classifiers.

\section{Discussion}
Although the SPSA optimizer improved robustness over other gradient-free methods, stochastic optimization noise and limited circuit expressivity constrained learning effectiveness in variational quantum classifiers. The results show clear differences between fully quantum models, hybrid quantum-classical models, and traditional machine-learning approaches for detecting coordinated stealth attacks in DG systems. Fully quantum variational classifiers consistently performed worse than classical models across all experiments. This behavior is primarily due to limitations of current NISQ-era quantum hardware, including noisy objective evaluations and barren plateau effects. As the circuit depth increased, training became unstable, and the model did not converge well, which agrees with earlier theoretical and experimental studies \cite{mcclean2018barren,zhang2022fundamental}. Shallow circuits were too simple to learn useful decision boundaries, while deeper circuits were difficult to optimize.

Hybrid quantum--classical models exhibited more reliable and effective behavior. By using quantum circuits exclusively for nonlinear feature embedding rather than end-to-end variational training, these models avoided unstable optimization and vanishing gradients. Quantum feature maps transformed electrical measurements into structured nonlinear representations that improved class separability. When combined with classical classifiers such as support vector machines, the hybrid models achieved the best overall performance. They slightly outperformed the classical SVM baseline while showing stable training and high attack detection accuracy \cite{schuld2019quantum,havlivcek2019supervised}. These results demonstrate that quantum embeddings can enhance intrusion detection performance even when fully quantum learning remains impractical.
\subsection*{Implications for Power-System Cybersecurity}
Although classical support vector machines achieved strong performance, this is expected because the feature space is low-dimensional. Classical models are well optimized and perform very well when only a small number of informative features are used, such as the three-feature space considered here. Quantum advantage is more likely to emerge in higher-dimensional or more complex learning tasks where classical kernel methods struggle \cite{biamonte2017quantum,ciliberto2018quantum}. In this context, quantum methods currently complement rather than replace classical intrusion detection techniques. Importantly, this work provides experimental evidence that quantum feature mappings can capture meaningful structure in real power-system measurements under coordinated stealth attack conditions, even when fully quantum learning is not yet practical \cite{eskandarpour2019quantum,ajagekar2019quantum}.

\section{Conclusion}
This paper evaluated quantum machine-learning methods for detecting coordinated stealth attacks in distributed generation systems. Using reactive power and frequency deviation measurements, comparisons were made between classical machine-learning models, fully quantum variational classifiers, and hybrid quantum–classical approaches. Classical support vector machines showed strong performance, reflecting their maturity and effectiveness on low-dimensional intrusion detection tasks. Fully quantum variational classifiers were less effective because they are difficult to train and are limited by barren plateau effects and current NISQ hardware constraints.  

In contrast, hybrid quantum–classical models trained more reliably and achieved the best overall performance by combining quantum feature embeddings with classical learning algorithms. Although the performance gains over classical models were modest, the results demonstrate that quantum feature mappings can enhance the representation of power-system measurements and improve detection robustness without requiring end-to-end quantum training. This work represents one of the first experimental studies applying quantum machine learning to coordinated stealth attacks in distributed generation units. As quantum hardware and algorithms continue to improve, hybrid and fully quantum learning models are expected to play a larger role in power-system cybersecurity, especially in higher-dimensional and more complex grid environments.

\section*{Acknowledgments}
Special thanks to Professor Brett McKinney for his guidance, support, and mentorship throughout the Quantum and Scientific Computing course.

\bibliographystyle{IEEEtran}   
\bibliography{references}      

\vfill

\end{document}